\documentclass{article} 
\usepackage{iclr2023_conference,times}

\usepackage{amsmath,amsfonts,bm}

\def\eqref#1{equation~\ref{#1}}

\def\1{\bm{1}}

\DeclareMathAlphabet{\mathsfit}{\encodingdefault}{\sfdefault}{m}{sl}
\SetMathAlphabet{\mathsfit}{bold}{\encodingdefault}{\sfdefault}{bx}{n}

\usepackage{hyperref}
\usepackage{url}
\usepackage{graphicx}
\usepackage{algorithm}
\usepackage[noend]{algpseudocode}
\usepackage{multirow}
\usepackage{wrapfig,lipsum,booktabs}
\title{Joint Attention-Driven Domain Fusion and Noise-Tolerant Learning for Multi-Source Domain Adaptation}

\author{Tong Xu, Lin Wang, Wu Ning, Chunyan Lyu, Kejun Wang, Chenhui Wang \\
Harbin Engineering University,
The Hong Kong university of Science and Technology \\
University of California, Los Angeles\\
}

\iclrfinalcopy 
\begin{document}

\maketitle

\begin{abstract}
Multi-source Unsupervised Domain Adaptation (MUDA) transfers knowledge from multiple source domains with labeled data to an unlabeled target domain. Recently, endeavours have been made in establishing connections among different domains to enable feature interaction. 
However, as these approaches essentially enhance category information, they lack the transfer of the domain-specific information. 
Moreover, few research has explored the connection between pseudo-label generation and the framework's learning capabilities, crucial for ensuring robust MUDA.
In this paper, we propose a novel framework, which significantly reduces the domain discrepancy and demonstrates new state-of-the-art performance. 
In particular, we first propose a Contrary Attention-based Domain Merge (CADM) module to 
enable the interaction among the features so as to achieve the mixture of domain-specific information instead of focusing on the category information.
Secondly, to enable the network to correct the pseudo labels during training, we propose an adaptive and reverse cross-entropy loss, which can adaptively impose constraints on the pseudo-label generation process. 
We conduct experiments on four benchmark datasets, showing that our approach can efficiently fuse all domains for MUDA while showing much better performance than the prior methods.
\end{abstract}

\section{INTRODUCTION}
Deep neural networks (DNNs) have achieved excellent performance on various vision tasks under the assumption that training and test data come from the same distribution. 
However, different scenes have different illumination, viewing angles, and styles, which may cause the domain shift problem~\citep{20,10,5}. 
This can eventually lead to a significant performance drop on the target task.

Unsupervised Domain Adaptation (UDA) aims at addressing this issue by transferring knowledge from the source domain to the unlabeled target domain \citep{3}.
Early research has mostly focused on Single-source Unsupervised Domain Adaptation (SUDA), which transfers knowledge from one source domain to the target domain. Accordingly, some methods align the feature distribution among source and target domains~\citep{4,8} while some \citep{10,11} learn domain invariants through adversarial learning. Moreover, some methods, \textit{e.g.}, \cite{46}, take the label information as an entry point to maintain the robust training process. 
However, data is usually collected from multiple domains in the real-world scenario, which arises a more practical task, i.e., Multi-source Unsupervised Domain Adaptation (MUDA) \citep{16}.

MUDA leverages all of the available data and thus enables performance gains; nonetheless, it introduces a new challenge of reducing domain shift between all source and target domains. For this, some research~\citep{19, 20} builds their methods based on SUDA, aiming to extract common domain-invariant features for all domains. 
Moreover, some works, \textit{e.g.}, \cite{21, DAEL} focus on the classifier's predictions to achieve domain alignment. Recently, some approaches \citep{54, DAGN} take advantage of the MUDA property to create connections for each domain. These methods attempt to achieve cross-domain consistency of category information by filtering domain-specific information. However, for multiple domains, filtering the domain-specific information for every domain is difficult and often results in losing discrimination ability. Therefore, sharing the same domain-specific information for all domains makes more sense, which can be achieved through transferring the domain-specific information based on the connections.
Moreover, existing methods ignore the importance of generating reliable pseudo labels as noisy pseudo labels could lead to the accumulation of prediction errors. Consequently, it is imperative to design a robust pseudo-label generation based on the MUDA framework.

\begin{figure}[tp]
	\begin{center}
		\includegraphics[width=1.0\linewidth]{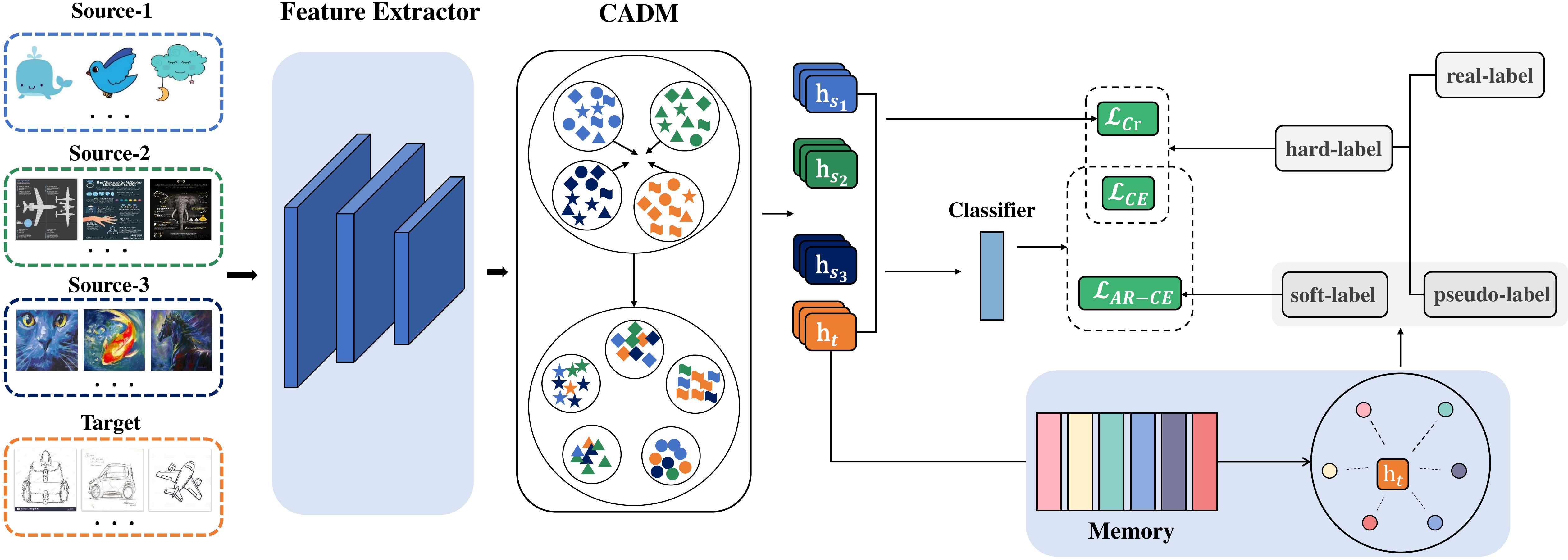}
	\end{center}
	\label{fig:Framework}
	\caption{
		Overview of our proposed framework that consists of three primary elements: (1) The feature extractor extracts features from various domains. (2) CADM is proposed to implement message passing and fuse features from different domains. The same domain is represented by the same color, while different shapes represent different classes. (3) we propose AR-CE loss and use the maintained memory to compute the soft label and pseudo label (hard label) for target domain.
	}
	\vspace{-3mm}
\end{figure}

In this paper, we propose a novel framework that better reduces the domain discrepancy and shows the new state-of-the-art (SoTA) MUDA performance, as shown in Fig.~\ref{fig:Framework}. Our method enjoys two pivotal technical breakthroughs. Firstly, we propose a Contrary Attention-based Domain Merge (CADM) module (Sec.~\ref{CADM}).
Cross attention can capture the higher-order correlation between features and emphases more relevant feature information, \textit{e.g.}, the semantically closest information.
Differently, our CADM module proposes the contrary cross-attention mechanism to focus on the less relevant domain-specific feature information and encourage its transfer. 
Then CADM is able to integrate the domain-specific information from other domains, thus achieving the deep fusion of all domains.
Secondly, To enable the network to correct the pseudo labels during training, we take the pseudo-label generation as the optimization objective of the network by proposing an adaptive and reverse cross-entropy (AR-CE) loss (Sec.~\ref{arce}). It imposes the optimizable constraints for pseudo-label generation, enabling the network to correct pseudo labels that tend to be wrong and reinforce pseudo labels that tend to be correct.

We conduct extensive experiments on four benchmark datasets. The results show our method achieves new state-of-the-art performance and especially has significant advantages in dealing with the large-scale dataset.
In summary, we have made the following contributions:
\begin{itemize}
	\item{We propose a CADM to fuse the features of source and target domains, bridging the gap between different domains and enhancing the discriminability of different categories.}
	\item{We propose a loss function, AR-CE loss, to diminish the negative impact of the noisy pseudo label during training.}
	\item{Our method demonstrates the new state-of-the-art (SoTA) MUDA performance on multiple benchmark datasets.}
\end{itemize}

\section{RELATED WORK}
\label{sec: Related Work}
\subsection{Unsupervised Domain Adaptation (UDA)}
\label{UDA}
According to the number of source domains, UDA can be divided into Single-source Unsupervised Domain Adaptation (SUDA) and Multi-source Unsupervised Domain Adaptation (MUDA). In SUDA, some methods use the metrics, such as Maximum Mean Discrepancy \citep{7, 26}, to define and optimize domain shift. Some methods \citep{31, GCAN} learn the domain-invariant representations through adversarial learning. Other methods are based on the label information. \cite{46} aligns domains implicitly by constraining the distribution of prediction. Following \cite{46}, \cite{SHOT++} introduces a label shifting strategy as a way to improve the accuracy of low confidence predictions. 
For MUDA, previous methods eliminate the domain shift while aligning all source domains by constructing multiple source-target pairs, where \citep{19, 20} are based on the discrepancy metric among different distributions and \citep{21, ptmda} are based on the adversarial learning. Some methods \citep{21, most, dca, stem} explore the relation between different classifiers and develop different agreements to achieve domain alignment. Some \citep{DAGN, 51} achieve an adaptive transfer by a weighted combination of source domains. Current methods \citep{LTC, 54} mainly construct connections between different domains to enable feature interaction and explore the relation of category information from different domains. Our approach also constructs connections between different domains; however, it mainly transfers domain-specific information to achieve the fusion of all domains.

\subsection{Robust Loss Function under Noisy Labels}
A robust loss function is critical for UDA because the unlabeled target domain requires pseudo labels, which are often noisy. Previous work \citep{23} demonstrates that some loss functions such as Mean Absolute Error (MAE) are more robust to noisy labels than the commonly used loss functions such as Cross Entropy (CE) loss. \citep{25} proposes the Symmetric Cross Entropy (SCE) combining the Reverse Cross Entropy (RCE) with the CE loss. Moreover, \citep{38} shows that directly adjusting the update process of the loss by the weight variance is effective. Some methods \citep{21, DAEL} mentioned in Section~\ref{UDA} focus on exploring a learning strategy to handle prediction values that have high or low confidence levels, respectively. Differently, we design an adaptive loss function for the network to learn robust pseudo-label generation by self-correction.

\section{PROBLEM SETTING}
\label{sec: Problem Setting}
\def\*#1{\mathbf{#1}}
\def\^#1{\mathnormal{#1}}
For MUDA task, there are $N$ source distributions and one target distribution, which can be denoted as $\{p_{s_j}(x,y)\}_{j=1}^N$ and $\{p_t(x,y)\}$. The labeled source domain images $\{X_{s_j}, Y_{s_j}\}_{j=1}^{N}$ are obtained from the source distributions, where $\mathcal X_{s_j}=\{{x^{i}_{s_j}}\}_{i=1}^{\lvert \mathcal X_{s_j} \rvert}$ are the images in the source domain $j$ and $\mathcal Y_{s_j} = \{{y^{i}_{s_j}}\}_{i=1}^{\lvert \mathcal X_{s_j} \rvert}$ represents the corresponding ground-truth labels. As for the unlabeled data in the target domain, there are target images $\mathcal X_{t}=\{{x^{i}_{t}}\}_{i=1}^{\lvert \mathcal X_{t} \rvert}$ from target distribution.
In this paper, we uniformly set the number of samples in a batch as $B$, with $b = \frac{B}{N + 1}$ for every domain (including the target domain).
For a single sample $x_{s_j}^{i}$, the subscript $s_j$ represents the $j$-th source domain, while the superscript $i$ indicates that it is the $i$-th sample in that domain.
\section{PROPOSED METHOD}
\subsection{Overall Framework}
The general MUDA pipeline consists of two phases: 1) pre-training on the source domain; 2) training when transferring to the target domain. The base structure of the model consists of a feature extractor $F(\cdot)$ and a classifier $C(\cdot)$.
In the pre-training phase, we use the source domain data $\{X_{s_j}, Y_{s_j}\}_{j=1}^{N}$ for training to obtain the pre-trained $F(\cdot)$ and $C(\cdot)$.
In the training phase for the target domain, we firstly feed the source domain data $\{X_{s_j}, Y_{s_j}\}_{j=1}^{N}$ and target domain data  $X_{t}$ into $F(\cdot)$ to get the features $\mathbf f$, and perform the feature fusion by CADM to get the fused feature $\* h$. Then, we use the fused features of the target domain for the pseudo-label generation and feed the fused features of all domains into $C(\cdot)$ to get prediction values. Finally, we compute the loss based on the labels. Since source training is general, training in this paper typically refers to target domain training.
\subsection{Contrary Attention-based Domain Merge Module}
\label{CADM}
\begin{figure*}[t]
	\centering
	\setlength{\abovecaptionskip}{0.1cm}
	\includegraphics[width=1.0\linewidth]{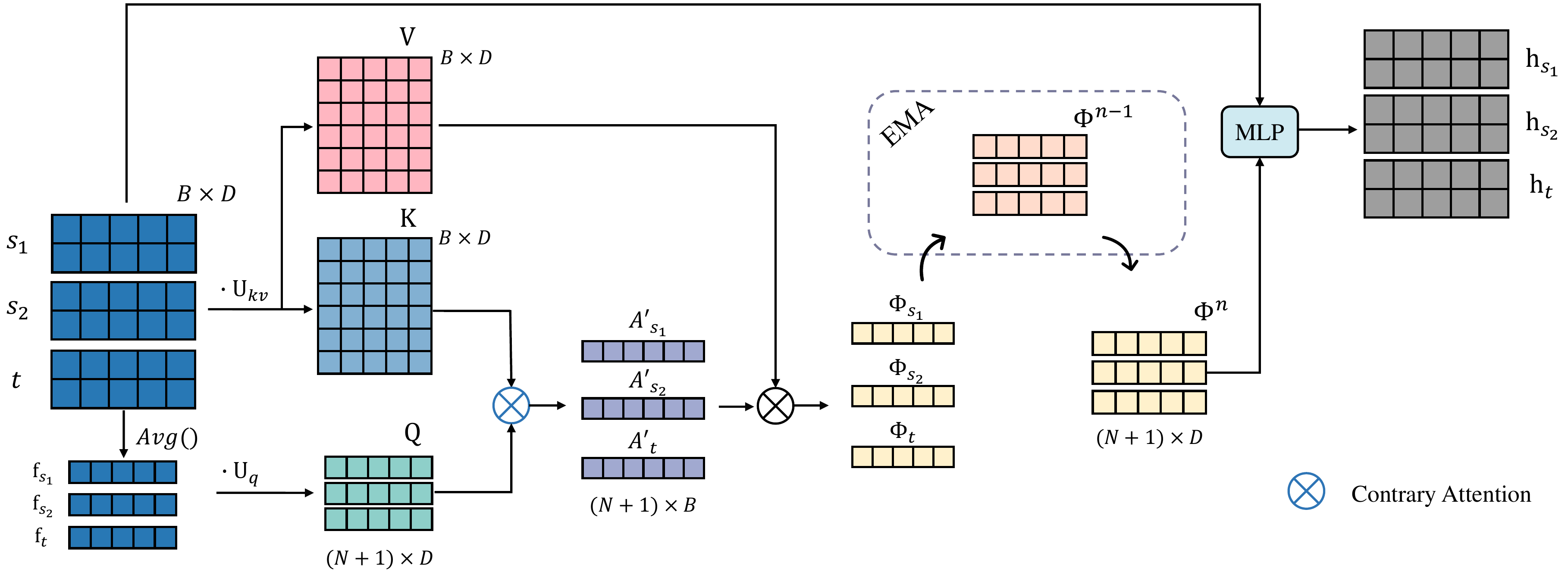}
	\caption{CADM review: Features from different domains are fused by CADM. For simplicity, we set $N$, $B$, and $D$ as 2, 6, and 5, respectively. Domain prototypes are mapped to $\mathbf Q$, while all features are mapped to $\mathbf K$ and $\mathbf V$. The domain style centroids are obtained by the contrary attention map $A'$ and $\mathbf V$, which is then updated by EMA. Finally, the fused feature $\* h$ is obtained by MLP.}
	\label{fig:3}
	\vspace{-0.4cm}
\end{figure*}
The proposed CADM enables the passing of domain-specific information to achieve the deep fusion of all domains. \textit{The main idea is to allow features to receive information from other domains in an adaptive manner based on their relevance}. The specific structure is shown in Fig.~\ref{fig:3}.
First, we use each domain to query the correlation with other domains. Specifically, samples are first mapped into the latent space via $F$ to obtain feature $\mathbf f$. Then, we define the mean embedding of features in a domain as the domain prototype:
\begin{equation}
	\label{eq0}
	\mathbf f_g = \dfrac{1}{b}\ \sum_{i=1}^{b}\ \mathbf f_{g}^{i}
\end{equation}
where $\mathbf f_g$ represents the domain prototype of domain $g$ and is then used to compute correlations with features in other domains as shown below:
\begin{align}
	\label{eq1}
	\mathbf {Q} &= [\* {f}_{s_1} \ldots \*{f}_{s_N}, \* {f}_{t}] \*U_{q} \hspace{2em} &\mathbf{U}_{q} &\in \mathbb{R}^{D\times D}\\
	\label{eq2}
	[\mathbf{K, V}] &= {[\mathbf f^1_{s_1} \ldots \mathbf f^1_{s_2}\ldots \mathbf f^b_{t}] \*U}_{kv} \hspace{2em} &\mathbf{U}_{kv} &\in \mathbb{R}^{D\times2D}\\
	\label{eq3}
	A_{g} &= \mathrm{Softmax}\left(\frac{\mathbf q_{g} \cdot \mathbf K^\mathrm{T}}{\sqrt{D}}\right) \hspace{2em} &A_{g} &\in \mathbb{R}^{1\times B}
\end{align}
where $\mathbf{U}_{q}$ and $\mathbf{U}_{kv}$ are learnable projection matrices that are used to map the domain prototype and all features to the new space respectively. We use the cross-attention mechanism \citep{40, 41} to model the correlation because of its powerful ability to capture dependencies. In Eq.~\ref{eq3}, $\* q_g$ is the query feature of the domain $g$, and $A_g$ is the obtained attention map through cross attention.

With the above formula, more emphasis is given to features that are more semantically close to the domain prototype, i.e., domain styles similar to $g$. However, such emphasis primarily strengthens the original domain style information of $g$ and does not help eliminate the domain shift. Instead, we take a different perspective, forcing the network to focus on features that are semantically distinct from the domain prototypes, which often have different domain-specific information. Specifically, we `reverse' the obtained $A_g$ to generate a new attention map, which we refer to as the contrary attention map:
\begin{equation}
	\label{eq4}
	A'_{g} = \frac{\* 1 - A_g}{\mathrm{sum}(\* 1 - A_g)} \hspace{3em} A'_{g} \in \mathbb{R}^{1\times B}
\end{equation}
where $\* 1$ is an all-one vector. The $A'_g$ thus obtained can achieve the opposite effect of the previous $A_g$, where those features that are more different from the domain $g$ in terms of domain-specific information are assigned higher weights.

Subsequently, we use $A'_g$ to perform the feature fusion for domain $g$, which allows the merging of information from features according to their degree of dissimilarity with domain $g$.
{\begin{equation}
		\begin{split}
			\label{eq5}
			\Phi_g &= A'_{g} \cdot \mathbf{V}\\
			&=\displaystyle\sum_{\hat g\in \mathcal {G}}\ \displaystyle\sum_{i=1}^{b} a'_{g\hat g^i} \mathbf{v}_{g}^i
		\end{split}
\end{equation}}
where $a'_{g\hat g^i}$ represents the contrary attention weight between the prototype of $g$ and the $i$-th feature in $\hat g$. $\Phi_g$ is obtained by weighting features from different domains and summing them according to their discrepancies from $g$. Given its greater emphasis on the merging of knowledge with distinct styles, we call $\Phi_g$ the domain style centroid with respect to $g$. 
In addition, to guarantee the robustness of $\Phi_g$, we use exponential moving averages (EMA) to maintain it:
\begin{equation}
	\label{eq6}
	(\bar{\Phi_g})^n = (1 - \alpha)(\bar{\Phi_g})^{n-1} + \alpha (\Phi_g)^n
\end{equation}
where $n$ denotes the $n$-th mini-batch. The recorded $\bar{\Phi_g}$ will be used in the inference phase.

Finally, we integrate the domain style centroid $\bar {\Phi_g}$ into the original features. A fundamental MLP is used to simulate this process:
\begin{equation}
	\label{eq7}
	\mathbf h^i_g = \mathrm {MLP}\left([\mathbf f_{g}^{i}\ , \ \bar{\Phi_g}]\right)
\end{equation}
where $\mathbf h^i_g$ is the fused feature and $\bar{\Phi_g}$ is integrated into every feature in $g$, thus creating the overall movement of domain. Through the process described above, the original features achieve the merging of different domain-specific information by integrating the domain style centroid. Meanwhile, to maintain a well-defined decision boundary during such domain fusion, we use a metric-based loss \citep{center} to optimize the intra-class distance. It allows samples from the same categories to be close to each other and shrink around the class center.
\begin{equation}
	\label{eq9}
	\mathcal{L}_{Cr} = \displaystyle {\sum_{g\in \mathcal{G}} \sum_{i=1}^b \ \Vert \mathbf h_{g}^{i} - \mu_{y_{g}^{i}}\Vert_2^2}
\end{equation}
where $y_{g}^{i}$ is the label of $\mathbf h_{g}^{i}$ and $\mu_{y_g}^i \in \mathbb{R}^{1\times D}$ denotes the $y_g^i$-th class center. For the robustness of class center, we maintain a memory $\mathrm{M}$ to store features at the class level and update them batch by batch during training, which is widely applied in UDA \citep{memory}. The class-level features can be regarded as the class center, and we then apply them in the computation of $\mathcal L_{Cr}$. Specifically, we traverse all the datasets to extract features and obtain different class centers before the training process. In the target domain, we use the classification result as the category since there is no ground-truth label:
\begin{equation}
	\label{eq10}
	\mu_{k} = \mathrm{Avg}\Big({\sum_{j=1}^{N} \displaystyle \sum_{x_i\in \mathcal X_{s_j}}^{y_i=k} \* h_{s_j}^i + \displaystyle \sum_{x_i\in \mathcal X_{t}}^{\arg\max\limits_{j} p_j(x_i)=k} \* h_{t}^i}\Big)
\end{equation}
where $p_j(x)$ represents the confidence level of predicting $x$ as the $j$-th class.
Then after each backpropagation step, we update the memory $\mathrm{M}$ with the following rule:
\begin{equation}
	\label{eq11}
	\mu_{k}^{new}=(1 - \beta) \mu_{k}^{old} + \beta \mathbf h
\end{equation}
where $k$ is the label of $\mathbf h$ and $\mu_{k}$ is the $k$-th class center in $\mathrm M$.
Under the supervision of $\mathcal L_{Cr}$, the features are able to maintain intra-class compactness in domain movement.

Overall, through CADM, we establish connections among domains and enable the deep fusion of different domains. Eventually, with the $\mathcal L_{Cr}$, the model is able to reach a tradeoff between domain merging and a well-defined decision boundary.
\subsection{Adaptive and Reverse Cross Entropy Loss}
\label{arce}
General UDA methods use the K-means-like algorithm to generate pseudo labels for the entire unlabeled target dataset. However, it has been observed that the pseudo labels generated in this way are often misassigned due to the lack of constraints, which introduces noise and impairs performance.
To address this issue, we creatively propose to assign gradients to the generation process of pseudo labels and constrain it with a well-designed rule, so that the network can optimize both the prediction values and pseudo labels simultaneously.
To begin, we expect to obtain the pseudo labels with gradient flow. Since the dynamically updated memory $\mathrm M$ is introduced in Section~\ref{CADM}, we can obtain both soft labels and hard labels by computing the distance between features and the class-level features in $\mathrm M$ in every batch:
\begin{align}
	\label{eq12}
	q(y=k\vert x) &=\dfrac {e^{\mathbf h \cdot \mu_{k}^{\mathrm{T}}}} {\displaystyle\sum_{j=1}^{K}e^{\mathbf h \cdot \mu_{j}^{\mathrm{T}}}}\\
	\label{eq13}
	\hat y &= \arg \max \limits_{k} q(y=k\vert x)
\end{align}
where $q(y\vert x)$ is the soft label and $\hat y$ is the hard label. The soft label thus obtained makes the gradient backpropagation possible and can be supervised by a designed loss function. Furthermore, due to the noise in pseudo labels, this generation may lead to error accumulation and is therefore required to be constrained. Previous experimental observations \citep{21, 25} demonstrate that the trained classifier has abundant information, which is key to preventing the model from collapsing in the unlabeled target domain. Therefore, based on the prediction values of the classifier, we establish constraints on the pseudo-label generation by a reverse cross entropy loss:
\begin{equation}
\label{eqrce}
    \ell_{rce} = \displaystyle \sum_{k=1}^{K}p_{k}(x)\log q(y=k\vert x)
\end{equation}
where $p_k(x)$ represents the confidence level of predicting $x$ as the $k$-th class. Such a constraint prevents the overly aggressive pseudo-label generation and error accumulation. However, it is unfair to apply the same rule to those pseudo labels that tend to be accurate. An ideal situation is that the constraints can be adaptively adjusted to correspond to different pseudo labels.
Since pseudo label essentially arises from the feature extracted by the network, its accuracy is closely related to the ability of the network to distinguish this sample. \cite{SHOT++} demonstrates that when the model has a good distinguishing ability for one sample, the corresponding prediction values should be close to the one-hot vector. Inspired by it, we design an adaptive and reverse cross entropy loss:
\begin{align}
	\label{dis}
	&\hspace{5em} \delta_{j}(x, \hat y) = \frac{p_j(x)}{1 - p_{\hat y}(x)}\\
	\label{eq14}
	& \ell_{arce} (x, \hat y) = 
	\dfrac{{\ell_{rce}}}{\exp\bigg(\Big(-{\displaystyle \sum_{j}^{j\neq \hat y}\delta_{j}(x, \hat y)\log \delta_{j}(x, \hat y)}\Big)\ \big{/}\ \tau\bigg)}\\
	\label{eq15}
	&\hspace{5em} \mathcal L_{AR-CE} = \displaystyle\sum_{i}^{B} \ell_{arce}(x_i, \hat y_i)
\end{align}
where $\hat y$ is the generated hard label of the target domain sample $x$.
$\delta_j(x,\hat y)$ in Eq.~\ref{eq14} is the distribution of the prediction values except for the prediction values of the class $\hat y$ (hard label).
The low entropy of this distribution represents the existence of a high prediction value except for the class $\hat y$, which is not in accordance with the previously mentioned one-hot vector assumption, and therefore the network has a relatively weak distinguishing ability for this sample. Similarly, a high entropy of this distribution represents the network tends to make the correct prediction and has a good distinguishing ability for this sample.
As a result, we use the entropy of this distribution to implement an adaptive adjustment mechanism for the constraints, as shown in Eq.~\ref{eq14}. The exponential form and $\tau$ are used to amplify its effect. Since $\mathcal L_{AR-CE}$ aims to focus on the optimization of pseudo label, we freeze the gradient of the prediction values in Eq.~\ref{eqrce}. The backpropagation of $\mathcal L_{AR-CE}$ can eventually enable the optimization of pseudo-label generation.

Furthermore, the usual cross entropy loss is denoted as follows:
\begin{equation}
	\mathcal L_{CE} = -\frac{1}{B} \sum_{i=1}^{B}\log p_{y_i}(x_i)
	\label{eq17}
\end{equation}
where $y_i$ is the ground-truth label for source domain sample and is the hard pseudo label $\hat y_i$ obtained by Eq.~\ref{eq13} when $x_i$ is in the target domain.

Ultimately, the loss function for the entire framework can be represented as follows:
\begin{equation}
	\label{eq18}
	\mathcal L = \mathcal L_{CE}  + \mathcal L_{Cr} + \mathcal L_{AR-CE}
\end{equation}
where $\mathcal L_{CE}$ is the normal cross entropy loss, $\mathcal L_{Cr}$ is for feature fusion, $\mathcal L_{AR-CE}$ is used for noise tolerant learning, and the latter one item acts only on the target domain.

\section{Experiments}
\subsection{Datasets and Implementation}
Office-31 \citep{3} is a fundamental domain adaptation dataset comprised of three separate domains: Amazon (A), Webcam (W), and DSLR (D). 
Office-Caltech \citep{42} has an additional domain Caltech-256 (C) than Office-31. Office-Home \citep{43} is a medium-size dataset which contains four domains: Art (Ar), Clipart (Cl), Product (Pr), and Real-World (Rw). DomainNet \citep{19} is the largest dataset available for MUDA. It contains 6 different domains:
Clipart (Clp), Infograph (Inf), Painting (Pnt), Quickdraw (Qdr), Real (Rel), and Sketch (Skt).

For a comparison with our method, we introduce the SoTA methods on MUDA, including DCTN \citep{18}, $ \text{M}^{3}\text{SDA}$ \citep{19}, MFSAN \citep{20}, MDDA \citep{MDDA}, DARN \citep{DAGN}, Ltc-MSDA \citep{LTC}, SHOT++ \citep{SHOT++}, SImpAl \citep{21}, T-SVDNet \citep{54}, CMSS \citep{51}, DECISION \citep{51}, DAEL \citep{DAEL}, PTMDA \citep{ptmda} and DINE \citep{dine}.

For Office-31, Office-Caltech and Office-Home, we use an Imagenet \citep{49} pre-trained ResNet-50 \citep{45} as the feature extractor. As for the DomainNet, we use ResNet-101 instead. In addition, a bottleneck layer (consists of two fully-connected layers) is used as the classifier. We use the Adam optimizer with a learning rate of $1 \times 10^{-6}$ and a weight decay of $5 \times 10^{-4}$. We set the learning rate of extractor to ten times that of the classifier. The hyperparameters $\alpha$ and $\beta$ are set to 0.05 and 0.005, respectively. All the framework is based on the Pytorch \citep{47}.
\subsection{Results}
We compare the ADNT with different approaches on four datasets, and the results are shown in the following tables. Note that the one pointed by the arrow is the target domain and the others are the source domains.
\begin{table*}[t] \centering
\begin{minipage}[t]{0.48\linewidth}
    \caption{Classification accuracies (\%) on Office-31 dataset, `` * " indicates that the method is based on ResNet-101.} \label{table:1}
    \resizebox{\textwidth}{!}{
    \renewcommand{\arraystretch}{1.1}
		\begin{tabular}{lcccc}
			\hline\noalign{\smallskip}
			\multirow{2}{*}{Method} & A,W & A,D & D,W & \multirow{2}{*}{Avg} \\
			 & $\rightarrow$D & $\rightarrow$W & $\rightarrow$A & \\
            \noalign{\smallskip}
			\hline
			\noalign{\smallskip}
			MDDA & 99.2 & 97.1 & 56.2 & 84.2\\
			LtC-MSDA & 99.6 & 97.2 & 56.9 & 84.6\\
			DCTN & 99.3 & 98.2 & 64.2 & 87.2\\
			MFSAN & 99.5 & 98.5 & 72.7 & 90.2\\
			$\text {SImpAl}_{50}$ & 99.2 & 97.4 & 70.6 & 89.0\\
			DECISION & 99.6 & 98.4 & 75.4 & 91.1\\
			DINE* & 99.2 & 98.4 & \bf 76.8 & 91.5\\
			PTMDA & \bf 100 & \bf 99.6 & 75.4 & \bf 91.7\\
			\bf ADNT (Ours) & \bf 100 & \bf 99.6 & 74.4 & 91.3\\
			\hline
		\end{tabular}
    }
\end{minipage}\hfill
    \begin{minipage}[t]{0.50\linewidth}
    \caption{Classification accuracies (\%) on Office-Caltech dataset. `` * " indicates that the method is based on ResNet-101.} \label{table:2}
    \resizebox{\textwidth}{!}{
    \renewcommand{\arraystretch}{1.26}
    \setlength\tabcolsep{3pt}
    \begin{tabular}{lccccc}
		    \hline\noalign{\smallskip}
			\multirow{2}{*}{Method} & A,C,D & A,C,W & A,W,D & C,D,W & \multirow{2}{*}{Avg} \\
			  & $\rightarrow$W & $\rightarrow$D & $\rightarrow$C & $\rightarrow$A &\\
			\noalign{\smallskip}
			\hline
			\noalign{\smallskip}
			DCTN & 99.4 & 99.0 & 90.2 & 92.7 &	95.3\\
			$\text{M}^{3}\text{SDA} $ & 99.4 &	99.2 &	91.5 &	94.1 &	96.1\\
			$\text {SImpAl}_{50}$ & 99.3 &	99.8 &	92.2 &	95.3 &	96.7\\
			CMSS & 99.3 & 99.6 & 96.6 & 93.7 & 97.2 \\
			SHOT++ & \bf 100 & 99.4 & 96.5 & 96.2 & 98.0 \\
			DECISION & 99.6 & \bf 100 & 95.9 & 95.9 & 98.0 \\
			DINE* & 98.9 & 98.5 & 95.2 & 95.9 & 97.1 \\
			PTMDA & \bf 100 & \bf 100 & 96.5 & \bf 96.7 & 98.3 \\
			\bf ADNT (Ours) & \bf 100 & \bf 100 & \bf 97.6 & 96.3 & \bf 98.5\\
			\hline
		\end{tabular}
    }
\end{minipage}
\end{table*}
\begin{table*}[!h]
	\small
	\begin{center}
		\caption{Classification accuracies (\%) on Office-Home dataset, `` * " indicates that the method is based on ResNet-101.}
		\label{table:3}
		\begin{tabular}{lccccc}
			\hline\noalign{\smallskip}
			\multirow{2}{*}{Method} & Cl,Pr,Rw & Ar,Pr,Rw & Ar,Cl,Rw & Ar,Cl,Pr & \multirow{2}{*}{Avg} \\
			& $\rightarrow$Ar & $\rightarrow$Cl & $\rightarrow$Pr & $\rightarrow$Rw & \\
			\noalign{\smallskip}
			\hline
			\noalign{\smallskip}
			MFSAN & 72.1 & 62.0 & 80.3 & 81.8 &	74.1\\
			$\text {SImpAl}_{50}$ & 70.8 & 56.3 & 80.2 & 81.5 &	72.2\\
			$\text {SImpAl}_{101}$* & 73.4 & 62.4 &	81.0 &	82.7 & 74.8 \\
			DARN & 70.0 & \bf 68.4 & 82.7 & 83.9 & 76.3 \\
			SHOT++ & 73.1 & 61.3 & 84.3 & \bf 84.0 & 75.7 \\
			DECISION & 74.5 & 59.4 & 84.4 & 83.6 & 75.5 \\
			DINE* & \bf 74.8 & 64.1 & 85.0 & \bf 84.6 & 77.1 \\
			\bf ADNT (Ours) & 73.8 & 66.5 & \bf 85.1 & 83.3 & \bf 77.2\\
			\hline
		\end{tabular}
	\end{center}
	\vspace{-2mm}
\end{table*}
Experimenting on the Office-31 dataset, we achieve optimal results in all scenarios, except for transferring W and D to A. On the Office-Caltech dataset, our method not only outperforms all current approaches, but also achieves an impressive $100\%$ accuracy on the task of transferring to W and to D.
In terms of the medium-sized Office-Home dataset, our proposed ADNT is the best performer and has a significant lead among the ResNet-50-based method. Compared with the ResNet-101-based methods (DINE and $\text {SImpAl}_{101}$), our approach achieves leading performance with fewer parameters and a simpler model structure.
As shown in Table~\ref{Table-DomainNet}, ADNT also demonstrates its strength on the largest and most challenging dataset. The total result outperforms the current optimal method by $2.3\%$, which is highly impressive on DomainNet.
\begin{table*}[h]
	\small
	\caption{Classification accuracies (\%) on DomainNet dataset.}
	\label{Table-DomainNet}
	\centering
	\begin{tabular}{lcccccccc}
		\toprule
		Methods    & $\rightarrow$ Clp & $\rightarrow$ Inf & $\rightarrow$ Pnt & $\rightarrow$ Qdr  & $\rightarrow$ Rel & $\rightarrow$ Skt &   Avg \\
		\midrule
		DCTN & 48.6$\pm$0.70 & 23.5$\pm$0.60 & 48.8$\pm$0.60 & 7.2$\pm$0.50 & 53.5$\pm$0.60 & 47.3$\pm$0.50 & 38.2 \\
		$ \text{M}^{3}\text{SDA}$ & 58.6$\pm$0.53 & 26.0$\pm$0.89 & 52.3$\pm$0.55 & 6.3$\pm$0.58 & 62.7$\pm$0.51 & 49.5$\pm$0.76 & 42.6  \\
		MDDA & 59.4$\pm$0.60 & 23.8$\pm$0.80 & 53.2$\pm$0.60 & 12.5$\pm$0.60 & 61.8$\pm$0.50 & 48.6$\pm$0.80 & 43.2 \\
		CMSS & 64.2$\pm$0.20 & 28.0$\pm$0.20 & 53.6$\pm$0.40 & 16.0$\pm$0.10 & 63.4$\pm$0.20 & 53.8$\pm$0.40 & 46.5 \\
		T-SVDNet & 66.1$\pm$0.4 & 25.0$\pm$0.8 & 54.3$\pm$0.7 & \bf{16.5$\pm$0.9} & 65.4$\pm$0.5 & 54.6$\pm$0.6 & 47.0\\
		LtC-MSDA & 63.1$\pm$0.5 & \bf{28.7$\pm$0.7} & 56.1$\pm$0.5 & 16.3$\pm$0.5 & 66.1$\pm$0.6 & 53.8$\pm$0.6 & 47.4\\
		DAEL & \bf{70.8$\pm$0.14} & 26.5$\pm$0.13 & 57.4$\pm$0.28 & 12.2$\pm$0.70 & 65.0$\pm$0.23 & 60.6$\pm$0.25 & 48.7  \\
		PTMDA & 66.0$\pm$0.3 & 28.5$\pm$0.2 & 58.4$\pm$0.4 & 13.0$\pm$0.5 & 63.0$\pm$0.24 & 54.1$\pm$0.3 & 47.2  \\
		\bf ADNT (ours) & 69.0$\pm$0.38 & 28.2$\pm$0.41 & \bf{60.5$\pm$0.45} & 16.3$\pm$0.66 & \bf{68.7$\pm$0.63} & \bf{63.5$\pm$0.69}& \bf{51.0} \\
		\bottomrule
	\end{tabular}
\end{table*}
\subsection{Ablation Study}
To verify the effectiveness of the proposed components, we conduct ablation experiments on the Office-Home dataset, as shown in Table~\ref{table:5}.
\begin{table*}[!t]
	\small
	\begin{center}
		\renewcommand{\tabcolsep}{0.8pc} 
		\renewcommand{\arraystretch}{1.1} 
		\caption{Classification accuracies (\%) of ablation study on Office-Home dataset.}
		\label{table:5}
		\begin{tabular}{lccccc}
			\toprule
			\multirow{2}{*}{Method} & Cl,Pr,Rw & Ar,Pr,Rw & Ar,Cl,Rw & Ar,Cl,Pr & \multirow{2}{*}{Avg} \\
			& $\rightarrow$Ar & $\rightarrow$Cl & $\rightarrow$Pr & $\rightarrow$Rw &\\
			\noalign{\smallskip}
			\hline
			\noalign{\smallskip}
			Baseline &	67.1 & 56.6 & 78.6 & 77.0 & 69.8\\
			+ $\mathcal{L}_{AR-CE}$ & 69.2 & 63.1 & 81.2 & 78.6 & 73.0\\
			+ ADM & 69.7 & 60.2 & 81.3 & 79.5 & 72.7\\
			+ CADM (w/o  $\mathcal L_{Cr}$) & 72.3 & 62.2 & 83.1 & 83.1 & 75.2\\
			+ CADM (w/  $\mathcal L_{Cr}$) & 73.1 & 64.2 & 84.3 & 83.7 & 76.3\\
			+ CADM (w/  $\mathcal L_{Cr}$) + $ \mathcal{L}_{AR-CE}$ & 73.8 & 66.5 & 85.1 & 83.3 & 77.2\\
			\hline
		\end{tabular}
	\end{center}
\end{table*}
In order to determine whether the attention-driven module actually achieves domain fusion, we conduct experiments with only the domain fusion without $\mathcal L_{Cr}$, which is shown as `+ CADM (w/o  $\mathcal L_{Cr}$)' in the table. Additionally, due to the fact that our proposed contrary attention differs from general attention by paying attention to the fusion of domain-specific information, a comparison with general attention can evaluate the viability of our strategy. Therefore, we replace the contrary attention map in the proposed CADM with a general attention map, i.e., without Eq.~\ref{eq4}. The result of general attention is shown as `+ ADM' in Table~\ref{table:5}. The significant performance improvement powerfully demonstrates the effectiveness of our proposed CADM.

\begin{figure*}[h]
	\centering
	\setlength{\abovecaptionskip}{0.cm}
	\includegraphics[width=\textwidth]{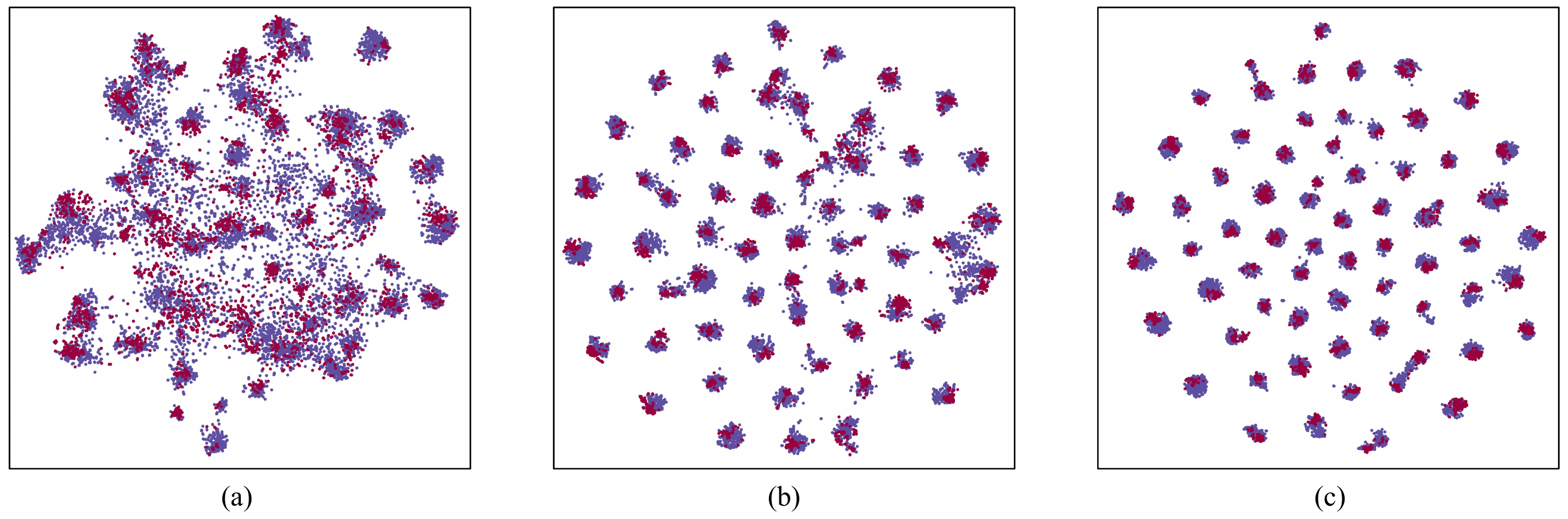}
	\caption{t-sne: The results of feature visualization on the Offce-Home dataset, while performing Ar, Cl, Rw to Pr. (a) (b) and (c) represent the result of Source-only, CADM without $\mathcal L_{Cr}$ and CADM with $\mathcal L_{Cr}$ respectively. All source domains are represented in red, while target domain is represented in blue.}
	\label{fig:4}
\end{figure*}

We also use t-sne \citep{tsne} to visualize the results with and without CADM, as shown in Fig.~\ref{fig:4}. We can observe that with the designed CADM, the different distributions are mixed with each other, achieving a higher degree of fusion. Meanwhile, the decision boundaries of different categories are more distinct.
\setlength{\tabcolsep}{4pt}
\begin{figure*}[t!]
	\centering
	\setlength{\abovecaptionskip}{0.cm}
	\includegraphics[width=\textwidth]{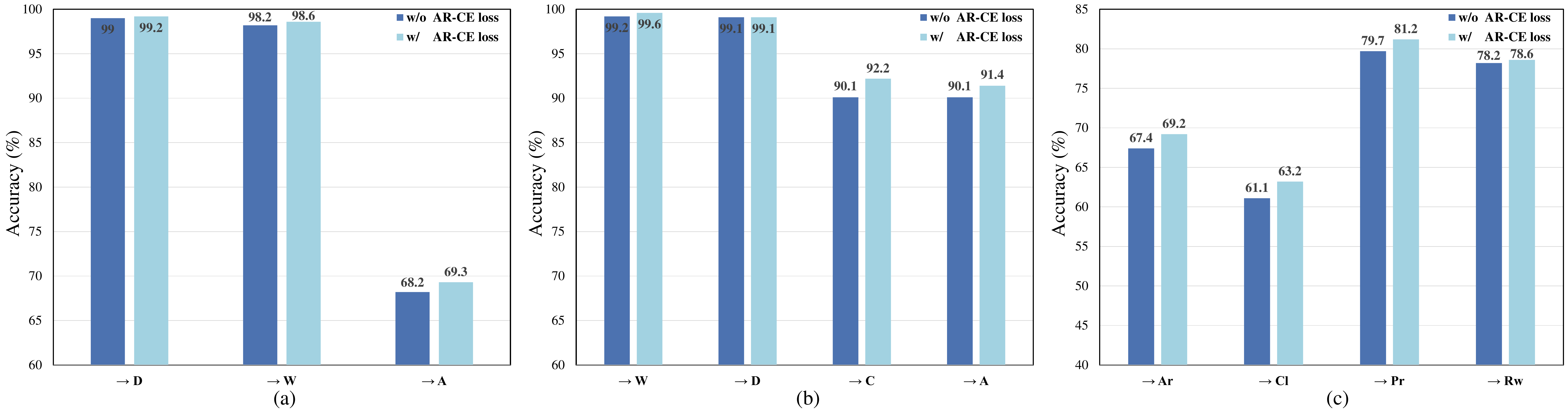}
	\caption{Mean accuracies (\%) of pseudo labels on three datasets. (a), (b), (c) represent the result on Office-31, Office-Caltech and Office-Home respectively.}
	\label{fig:5}
	\vspace{-0.4cm}
\end{figure*}
Table~\ref{table:5} verifies the effectiveness of our AR-CE loss. Moreover, the accuracy of the pseudo labels can reflect the effectiveness and stability of the pseudo-label generation. Therefore, we compare the average accuracy of pseudo labels generated with and without AR-CE loss on Office-31, Office-Caltech and Office-Home, which is shown in Fig.~\ref{fig:5}. Specifically, we calculate the mean accuracy of the final five epochs during training. It can be seen that after the introduction of the AR-CE loss, the model can derive noisy labels through self-correction and achieve the high accuracy of pseudo labels. Due to the space limit, additional analytical experiments, including the visualization of the training process and analysis of hyperparameters, can be found in the appendix.

\section{Conclusion}
This paper proposes the ADNT, a framework that combines the domain fusion module and robust loss function under noisy labels. Firstly, We construct a contrary attention-based domain merge module, which can achieve a mixture of different domain-specific information and thus fuse features from different domains. In addition, we design a robust loss function to avoid the error accumulation of pseudo-label generation. Substantial experiments on four datasets have shown the superior performance of ADNT, which strongly demonstrates the effectiveness of our approach.

\clearpage
\bibliography{iclr2023_conference}
\bibliographystyle{iclr2023_conference}

\clearpage
\appendix
\section{Appendix}

\subsubsection{Visualization of the Training}

In our proposed CADM, the weight value in $A'$ actually reflects the correlation between domains (Here, we use the domain to refer to the features in the domain). Therefore, We estimate the state between domains by analyzing $A'$.
In this paper, we expect the following domain relationship: Initially, they should have very different styles, and the weights in $A'$ will be quite uncertain. Then, as the optimization of the model parameters, different domains will gradually be mixed. Finally, all domains should have a consistent style and maintain a balanced relationship. Therefore, the merging of all domains together means that there should be little difference among the final correlation coefficients in $A'$. So we calculate the standard deviation of $A'$ during the training.
\begin{figure*}[!htbp]
	\centering
	\includegraphics[width=\textwidth]{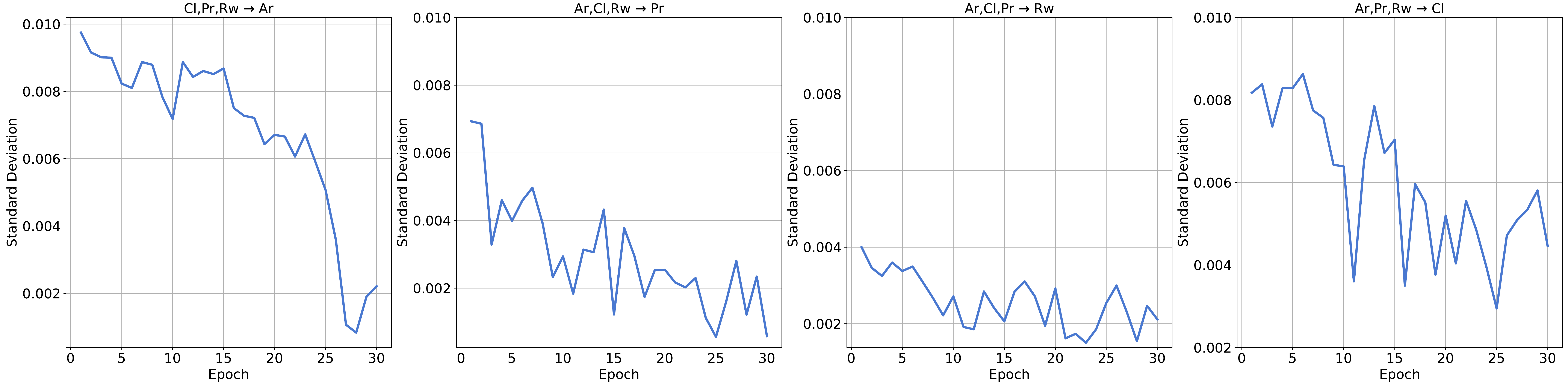}
	\caption{The standard deviation of $A'$ on the Office-Home dataset.}
	\label{fig:STD}
\end{figure*}
\begin{figure*}[!ht]
	\centering
	\includegraphics[width=1.0\linewidth]{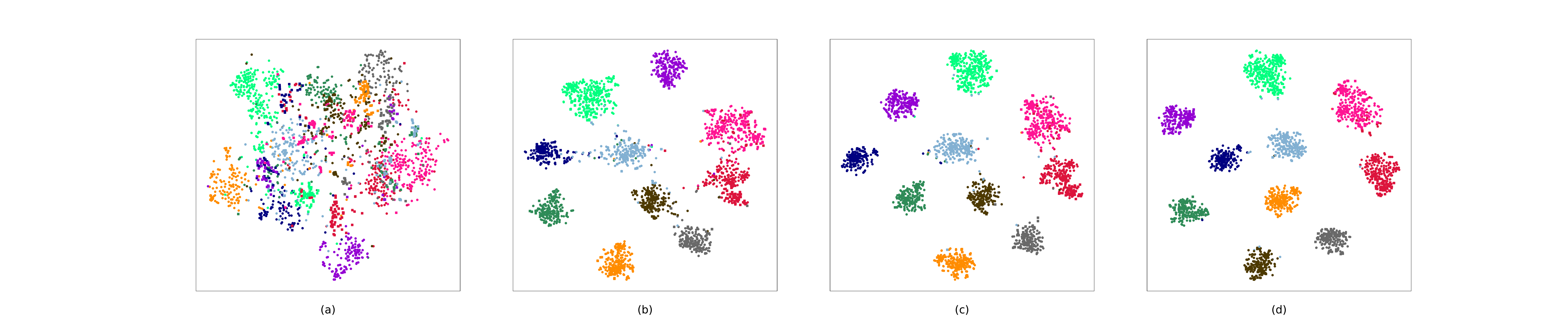}
	\caption{Visualization of the feature distribution during the training process in the transfer task Ar, Cl, Rw to Pr. (a), (b), (c) and (d) represent the epoch 1, 10, 20 and 30, respectively. To provide a clearer illustration, we select features from the first ten classes of each domain, with different color representing different class. Solid circles represent all source domains, while the squares represent target domains. Best viewed enlarged.}
	\label{fig:epoch}
\end{figure*}

As shown in the Fig.~\ref{fig:STD}, during the training process, the standard deviation gradually becomes smaller and eventually tends to be stable, which indicates that the features from different domains are distributed in the space with a uniform state.

However, the above results alone do not demonstrate that features from different domains are deeply fused because the previous effect in Fig.~\ref{fig:STD} can also be achieved if features are sufficiently separated from each other.
Consequently, we map the fused features $\mathbf{h}$ into the low-dimensional space and visualize the distribution during the training process, divided by different domains and categories. 
As shown in Fig.~\ref{fig:epoch}, at the start of training, the fused features are distributed haphazardly in space, and the network does not have capacity to process them. Then, different domains gradually move toward each other and merge together as the updating of model parameters. Finally, the features of the same category become more compact while preserving domain fusion. This powerfully demonstrates that through CADM, different domains embrace each other and achieve deep fusion. We make the model learn how to handle the fused features and show its excellent results.

\subsection{Parameter Analysis}
\begin{figure}[!htbp]
	\centering
	\includegraphics[width=1.0\linewidth]{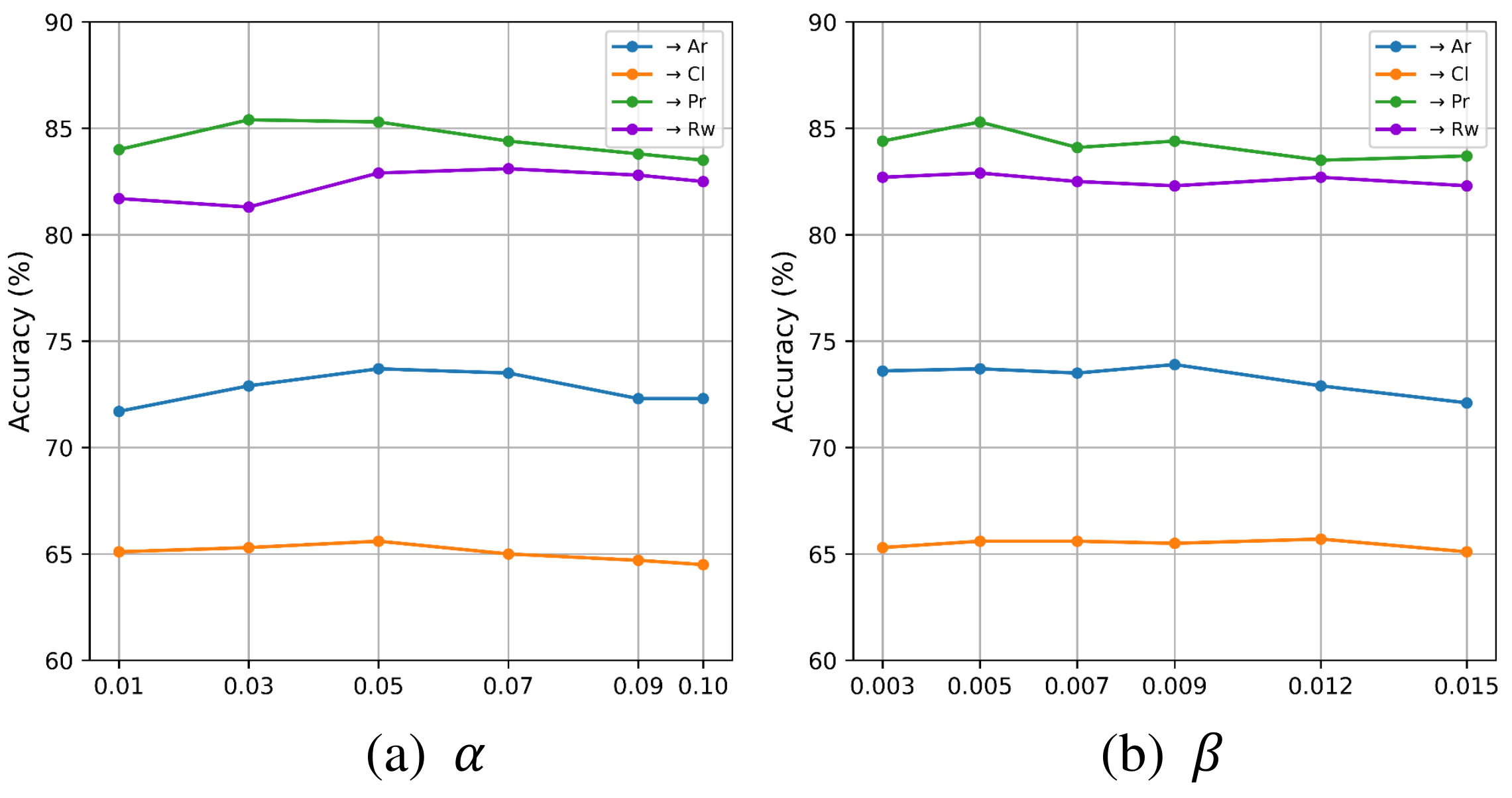}
	\caption{Sensitivity analysis of $\alpha$ and $\beta$.}
	\label{fig:param}
\end{figure}

In addition, exponential moving average (EMA) is used in the experiments to maintain the domain style centers and class-level features, so the selection of hyperparameters $\alpha$ and $\beta$ requires additional analysis and verification. Therefore, we perform an experimental comparison of their values and present them in Fig.~\ref{fig:param}.

As can be seen, the value of the hyperparameter $\alpha$ is somewhat greater and the $\beta$ is taken at an overall smaller value. This indicates that the domain style centroid is updated more vigorously and provides a gradient sufficient for the backpropagation of the network. Memory $\mathrm M$, on the other hand, requires a slower update to maintain network robustness because it involves the pseudo-label generation.

\end{document}